\documentclass{bmvc2k}
\usepackage{multirow}

\title{ASFormer: Transformer for Action Segmentation}

\addauthor{Fangqiu Yi}{chinayi@pku.edu.cn}{1}
\addauthor{Hongyu Wen}{1800013069@pku.edu.cn}{1}
\addauthor{Tingting Jiang}{ttjiang@pku.edu.cn}{1}

\addinstitution{
 NELVT\\
 Department of Computer Science\\
 Peking University, China
}

\runninghead{ASFormer}{Transformer for Action Segmentation}

\def\eg{\emph{e.g}\bmvaOneDot}

\def\etal{\emph{et al}\bmvaOneDot}

\begin{document}

\maketitle
\vspace{-0.5cm}
\begin{abstract}
Algorithms for the action segmentation task typically use temporal models to predict what action is occurring at each frame for a minute-long daily activity. Recent studies have shown the potential of Transformer in modeling the relations among elements in sequential data. However, there are several major concerns when directly applying the Transformer to the action segmentation task, such as the lack of inductive biases with small training sets, the deficit in processing long input sequence, and the limitation of the decoder architecture to utilize temporal relations among multiple action segments to refine the initial predictions. To address these concerns, we design an efficient Transformer-based model for action segmentation task, named ASFormer, with three distinctive characteristics: (i) We explicitly bring in the local connectivity inductive priors because of the high locality of features. It constrains the hypothesis space within a reliable scope, and is beneficial for the action segmentation task to learn a proper target function with small training sets. (ii) We apply a pre-defined hierarchical representation pattern that efficiently handles long input sequences. (iii) We carefully design the decoder to refine the initial predictions from the encoder.  Extensive experiments on three public datasets demonstrate that effectiveness of our methods. Code is available at \url{https://github.com/ChinaYi/ASFormer}.
\end{abstract}

\section{Introduction}
\label{sec:intro}
Algorithms for automatic detection and segmentation of human activities are crucial for applications such as home security, healthcare, and robot automatic. Different from the action classification task that tries to classify a short trimmed video into a single action label, the goal of the action segmentation task is to assign an action label for each frame for a minutes-long untrimmed video. Instead of using raw RGB video sequences as the input, action segmentation methods operate on pre-extracted frame-wise feature sequences and focus on modeling the temporal relations among frames. 
Transformers, originally designed for the machine translation task~\cite{att_all_you_need}, have achieved great performance for almost all natural language processing(NLP) tasks over the past years. Very recently, many researchers also show the potential of pure or hybrid Transformer models for many vision tasks, including image classification~\cite{transformer_cls1,transformer_cls2,transformer_cls3,transformer_cls4,transformer_cls5}, action classification~\cite{transformer_action1}, segmentation~\cite{transformer_seg1,transformer_seg2,transformer_seg3}, \etal. Action segmentation task is similar to NLP tasks, since both of them are sequence-to-sequence prediction tasks. With the success of Transformer-based models in modeling the relations among elements in sequential data, one would expect the Transformer-based models to be highly effective for the action segmentation task as well.

However, there are three major concerns when solving action segmentation task by the vanilla Transformer~\cite{att_all_you_need} :

\begin{enumerate}
    \item \textbf{Due to the small size of training sets, the lack of inductive biases} of the vanilla Transformer becomes the bottleneck of applying it to action segmentation problem. The lack of inductive biases broadens the family of functions they can represent~\cite{cordonnier2019relationship}, however, it requires a large amount of training data. Compared to the NLP task and other vision tasks, the training set of action segmentation task is relatively small, making it difficult to learn a target function from a large hypothesis space.
    
    \item \textbf{Due to the deficit of self-attention for the long input video, the Transformer is hard to form an effective representation}. At initialization, the self-attention layer cast nearly uniform attention weights to all the elements in the sequence. However, the input video for the action segmentation task usually lasts for thousands of frames, much longer than the image-patch sequences in other vision tasks. Due to the length of videos, it is challenging for the self-attention layer to learn appropriate weights that focus on meaningful locations~\cite{deformable_dert}. The deficit of each self-attention layer further raise a serious issue: it is hard for those self-attention layers in one Transformer model to cooperate with each other to form an effective representation for the input.
    
    \item \textbf{The original encoder-decoder architecture of the Transformer does not meet the refinement demand of action segmentation task.} Temporal relations among multiple action segments play an important role in the action segmentation task, \eg the action after \textit{take bottle} and \textit{pour water} usually to be \textit{drink water}. Given an initial prediction, previous works usually apply additional TCNs~\cite{MSTCN, BCN} or GCNs~\cite{GCN1,GCN2} over the initial prediction to perform a refinement process to boost the performance. However, the decoder in the vanilla encoder-decoder architecture is not designed for such usage.
\end{enumerate}

\begin{figure}[t]
	\centering
	\includegraphics[width=\textwidth]{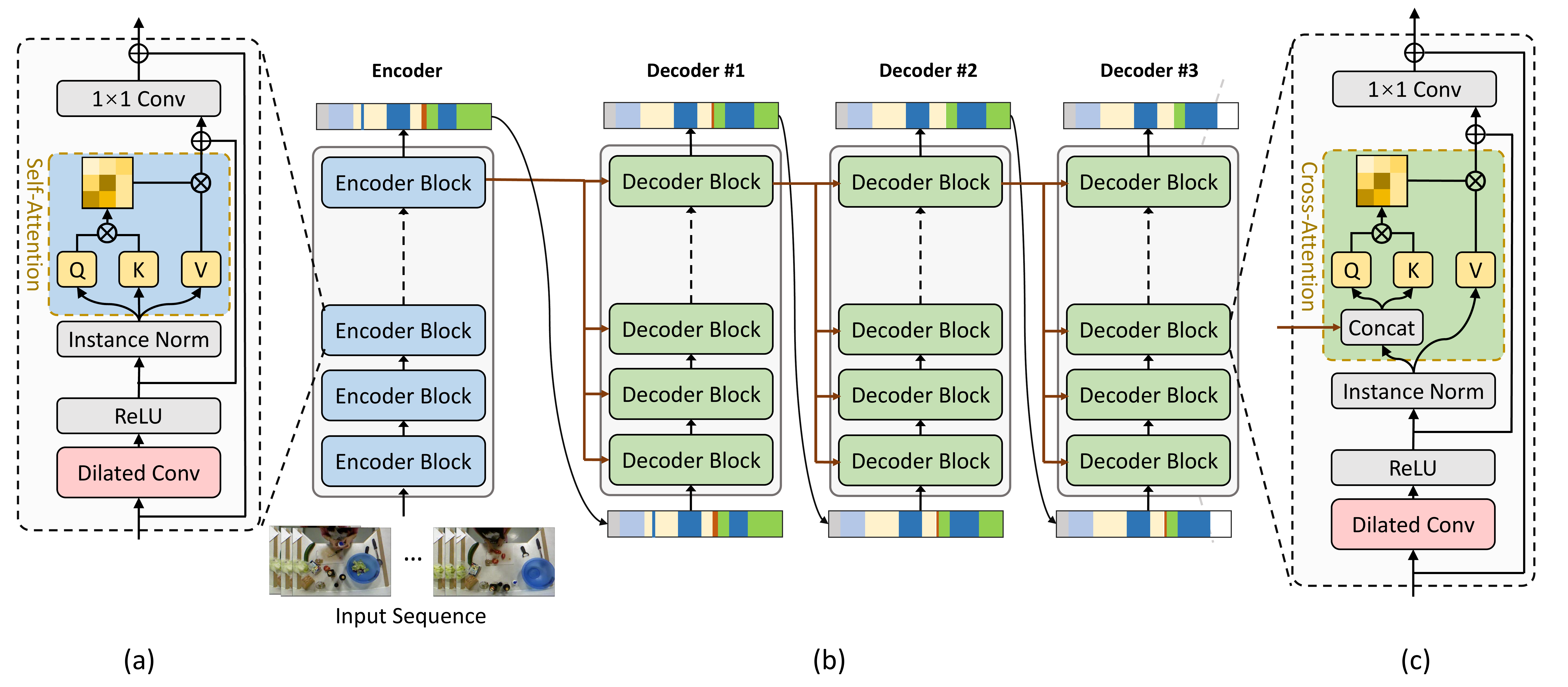}
	\caption{(b). An overall graph of the ASFormer model which consists of an encoder and several decoders to perform an iterative refinement. For the encoder, it receives video sequences and outputs initial predictions. Encoder consists of serials of encoder blocks with pre-defined hierarchical representation patterns. For the decoder, it receives predictions as the input and has similar architecture with encoder. (a). In each encoder block, it consists of a feed-forward layer (dilated temporal convolution) and a self-attention layer with residual connections.  (c). The decoder block uses cross-attention mechanism to bring in information from the encoder.}
	\label{workflow}
	\vspace{-0.3cm}
\end{figure}

In this paper, we will address the above three concerns in our proposed ASFormer, as shown in Fig.~\ref{workflow}. For the first concern, we observe that one property of the action segmentation task is the high locality of features because every action occupies continued timestamps. Thus, the \textbf{ local connectivity inductive bias} is important to the action segmentation task. It constrains the hypothesis space within a reliable scope, and is beneficial to learn a proper target function with small training sets. We bring in such strong inductive priors by applying additional temporal convolutions in each layer. For the second concern that the Transformer with serials of self-attention layers is hard to form an effective representation over the long input sequence, we constraint each self-attention layer with a \textbf{pre-defined hierarchical representation pattern}, which forces the low-level self-attention layers to focus on the local relations at first and then gradually enlarges their footprints to capture longer dependencies in high-level layers. The local-to-global process assigns specific responsibilities to each self-attention layer, so that they can cooperate better to achieve faster convergence speed and higher performance. Such hierarchical representation pattern also reduces the total space and time complexity to make our model scalable. Finally, we propose \textbf{a new design of the decoder} to acquire the refined predictions. The cross-attention mechanism in the decoder allows every position in the encoder to attend over all positions in the refinement process, simultaneously avoiding the disturbance of the encoder to the learned feature space in the refinement phase.

Experiments are conducted on three common public datasets, including the 50Salads~\cite{50salads}, Breakfast~\cite{breakfast} and GTEA~\cite{gtea}. The experimental results demonstrate the proposed solution is capable of dealing with small training dataset and long videos with thousands of frames. The design of decoders also takes advantage of temporal relations among multiple action segments to help get smoother and accurate predictions. To summarize, the main contributions of this work include: 1) An exploration for Transformer on action segmentation task with three distinctive characteristics: the explicitly introduced local connectivity inductive bias, pre-defined hierarchical representation pattern, and new design of the decoder; 2) state-of-the-art action segmentation results on three public datasets.

\section{Related Work}
\noindent
\textbf{Action Segmentation.} Earlier approaches detect action segments using a sliding window and filter out redundant hypotheses with non-maximum suppression~\cite{sliding1,sliding2}. Other approaches model the temporal action sequence with Conditional Random Fields~\cite{crf2,crf3,CRFs}, Markov model~\cite{HTK(64),hmm2} and RNN~\cite{lstm1,lstm2,lstm3,bi-lstm,lstm4} to classify framewise actions. In recent years, motivated by the success of temporal convolution in speech synthesis, most of the works try to apply various temporal convolution networks for action segmentation task, such as the encoder-decoder temporal convolution~\cite{ED-TCN,TCFPN}, deformable temporal convolution~\cite{TDRN}, or dilated temporal convolution~\cite{MSTCN,timeception}. Other works build on top of these TCNs. Wang \etal~\cite{BCN} reuse the structure of ~\cite{MSTCN} as cascade stages. Huang \etal~\cite{GCN1} and Wang \etal~\cite{GCN2} propose a graph-based temporal reasoning module, which can be added to the top of action segmentation models. Chen \etal~\cite{da1} propose to exploit auxiliary unlabeled videos by
shaping this problem as a domain adaptation (DA) problem. Very recently, Ishikawa1\etal~\cite{bound2} alleviate over-segmentation errors by detecting action boundaries. Singhania\etal~\cite{C2F} follow a coarse-to-fine structure with an implicit ensemble of multiple temporal resolutions. However, a single convolutional layer does not connect all pairs of input and output positions, which remains room for improvement. \\
\noindent
\textbf{Transformer.} Transformer~\cite{att_all_you_need} is originally designed for natural language processing tasks, which relies on the self-attention mechanism to build dependencies among elements. Due to the strong capability of capturing global information, hybrid Transformer models have also been successfully applied to a variety of vision problems, including image classification~\cite{transformer_cls1,transformer_cls2,transformer_cls3,transformer_cls4,transformer_cls5}, action classification~\cite{transformer_action1}, segmentation~\cite{transformer_seg1,transformer_seg2,transformer_seg3}, \etal. More recently, some researches study the efficient version of Transformer models, which explore attention restrictions to local windows, such as Swin~\cite{swin}, BigBird~\cite{bigbird}. In this paper, we explore the application of Transformer based models on the action segmentation task.


\noindent
\textbf{Action Detection.} Different from action segmentation task, the action detection task aims at localizing the start/end of the action and recognizing it in a untrimmed video. The One-stage methods~\cite{one_stage_action_det1,one_stage_action_det2} draw on the SSD~\cite{ssd} method in object detection and design end-to-end action detection networks with the similar feature pyramid structures. Two-stage methods~\cite{two_stage_action_det1,two_stage_action_det2} adopt the Faster-RCNN~\cite{faster_rcnn} architecture, including proposal generation and proposal classification. 


\section{Methods}
In this work, we propose ASFormer to tackle the action segmentation task, as shown in Fig.~\ref{workflow}. Our ASFormer adapts an encoder-decoder structured Transformer. Given the pre-extracted frame-wise video feature sequence, the encoder will first predict the initial action probability for each frame. Then the initial predictions will be passed to multiple successive decoders to perform an incremental refinement. In Sec.~\ref{encoder}, we first illustrate the structure of \textbf{encoder}, showing how we deal with small training dataset and long videos with thousands of frames. After that, we introduce the design of \textbf{decoders} and our way to take advantage of temporal relations among multiple action segments for refinement in Sec.~\ref{decoder}. Finally, in Sec.~\ref{details}, we introduce the implementation and training details of our ASFormer.

\subsection{Encoder}
\label{encoder}
The input for the encoder is the pre-extracted feature sequences of size $T \times D$, where $T$ is the video length and $D$ is the feature dimension. The first layer of the encoder is a fully connected layer that adjusts the dimension of the input feature. Then, this layer is followed by serials of encoder blocks. After that, a fully connected layer will output predictions $y_e \in {R}^{T \times C}$ from the last encoder block, where $C$ denotes the number of action classes.  

Each encoder block contains two sub-layers. The first is a feed-forward layer, and the second is a single-head self-attention layer. We employ a residual connection around each of the two sub-layers, followed by instance normalization and ReLU activation, as shown in Fig.~\ref{workflow}(a). Different from the vanilla Transformer, we use a dilated temporal convolution as the feed-forward layer instead of point-wise fully connected layer. This design is inspired by the properties of the action segmentation task, which are a) lack of the large training dataset for completely free inductive bias. b) high locality of features since every action occupies continued timestamps in the input video. In contrast to the fully connected layer, the temporal convolution layer can bring beneficial local inductive bias to our model.

The self-attention layer is hard to be learned to focus on meaningful locations over thousands of frames. For an input video, those self-attention layers are difficult to cooperate with each other to form an effective representation. To mitigate this issue, we pre-define a hierarchical representation pattern. Such a hierarchical pattern is inspired from modern neural network design: one would first focus on the local feature and then gradually enlarge the receptive field to capture the global information. For example, CNNs achieve such pattern with serials of pooling layers to enlarge the receptive field in higher layers; or use dilated convolutions with gradually increasing dilation rate. Motivated by the success of such a hierarchical pattern, we constraint the receptive fields of each self-attention layer within a local window with size $w$ (\eg{ for a frame $t$, we only calculate the attention weights with the frames that within its local window}). The size of the local window is then doubled at each layer $i$ (i.e., $w = 2^i, i=1,2...$). Meanwhile, we also double the dilation rate of the temporal convolution layer with the encoder depth increasing, keeping consistent with the self-attention layer.

For an encoder with ${J}$ blocks, the whole approximate memory usage for a vanilla Transformer is  $({J} \cdot T \cdot T)$, where $T$ is the video length. With the hierarchical representation pattern, We reduce the total space complexity to $((2-\epsilon)\cdot 2^{J} \cdot T)$, where $\epsilon$ is a small number. In our settings, we use ${J}=9$, where $2^{J}=512$ is almost 10 times smaller than $T$. Compared to the vanilla Transformer, our ASFormer is applicable to receiving long input sequence.

\subsection{Decoders}
\label{decoder}
Temporal relations among multiple action segments play an important role in the action segmentation task. There are some prior relationship between actions segments, \eg the action after \textit{take bottle} and \textit{pour water} usually to be \textit{drink water}. As previous works showed, applying additional TCNs~\cite{MSTCN, BCN} or GCNs~\cite{GCN1,GCN2} over the initial prediction to perform a refinement process can boost the performance. In this section, we illustrate how the new designed decoder carries out the refinement task for the initial predictions output by the encoder in one-forward-pass. For better explanation, we first introduce a single decoder, and naturally extend it to a multiple version to perform an iterative refinement. 

\noindent
\textbf{A Single Decoder.} 
The input for the decoder is the initial predictions output by the encoder. The first layer of the decoder is a fully connected layer to adjust the dimension, and then followed by a serial of decoder blocks. The architecture of each decoder block is shown in Fig.~\ref{workflow}(c). Similar to encoder, we use temporal convolution as the feed-forward layer and the hierarchical pattern is also applied for the cross-attention layer.

Compared to the self-attention layer, the cross-attention has the following difference: The query $Q$ and key $K$ are obtained from the concatenation of the output from the encoder and previous layer, while the value $V$ is only obtained from the output of the previous layer. The cross-attention mechanism allows every position in the encoder to attend over all positions in the refinement process by generating the attention weights. The feature space $V$ is completely transformed from the input predictions, and will not be disturbed with the participant of the encoder, because the generated attention weights are only used to perform a linear combination within $V$. Such design is inspired by the previous work~\cite{MSTCN}, where they show the refinement process is very sensitive to the disturbance of the learned feature space from the predictions. 

\textbf{Multiple Decoders.} One would naturally expand the single decoder to a multiple version to perform iterative refinement. In the multiple version, the input of each decoder are from the previous one, as shown in Fig.~\ref{workflow}(b). 

The cross-attention mechanism allows to bring in external information to guide the refinement process. We hope to gradually reduce the weight of external information to avoid the problem of error accumulation. For the input $x$ in each decoder block, we use a weighted residual connection for the output of the feed-forward layer and the cross-attention layer:
\begin{equation}
\begin{aligned}
  {out} &= {feed\_forward}(x) \\
  {out} &= \alpha * {cross\_att}({out}) + {out}
\end{aligned}
\end{equation}

We set the $\alpha=1$ for the first decoder and then exponentially decrease $\alpha$ for rest decoders.

\subsection{Loss Function \& Implementation details}
\label{details}
The loss function is a combination of classification loss $\mathcal{L}_{cls}$ for each frame and smooth loss $\mathcal{L}_{smo}$~\cite{MSTCN}. The classification loss is a cross-entropy loss, while the smooth loss calculates the mean squared error over the frame-wise probabilities. The final loss function $\mathcal{L}$ is,
$$\mathcal{L} = \mathcal{L}_{cls} + \lambda \mathcal{L}_{smo} = \frac{1}{T} \sum_{t}-log(y_{t,\hat{c}}) + \lambda \frac{1}{TC}\sum_{t}\sum_{c}(y_{t-1, c}-y_{t,c})^2$$ 
where $y_{t,\hat{c}}$ is the the predicted probability for the ground truth label $\hat{c}$ at time t. $\lambda$ is a balance weight that is set to 0.25 in our experiments. Finally to train the complete model, the sum of the losses over the encoder and all decoders is minimized.

The final ASFormer consists of one encoder and three decoders, while each encoder and the decoder contains nine blocks. The dimension of the first fully connected layer in the encoder and decoder is set to 64, as well as the feature dimension in each encoder and decoder block. Moreover, a special dropout layer is applied to the input feature of the encoder, which randomly drops the entire feature channel with a dropout rate of 0.3.  In all experiments, we train the model for 120 epochs through Adam optimizer with a learning rate 0.0005.

\section{Dataset}

\textbf{50Salads}~\cite{50salads} dataset contains 50 videos with 17 action classes of users making a salad and has been used for both action segmentation and detection. On average, each video contains 20 action instances and is about 6.4 minutes long. These activities were performed by 25 actors where each actor prepared two different salads. For evaluation, we perform 5-fold cross-validation and report the average results.

\noindent
\textbf{GTEA}~\cite{gtea} dataset contains 28 videos of 11 action classes of daily activities in a kitchen, like take or pour, performed by four subjects. On average, each video has 20 action instances and is about half-minute long on average. We use the standard four different train-test splits as previous works by leaving one subject out. Similar to 50Salads, the average results of four splits are reported.

\noindent
\textbf{Breakfast}~\cite{breakfast} dataset is the largest and the most challenging dataset among the three datasets with 1712 videos. The videos were recorded in 18 different kitchens, showing breakfast preparation-related activities. Overall, there are 48 different actions, and each video contains six action instances on average. We use the standard 4-fold cross-validation for evaluation and report the average results.

\noindent
For all the three datasets, we use the I3D~\cite{kinetics:i3d} model, which is trained on kinetics~\cite{kinetics:i3d} dataset, to pre-extract feature sequences as in previous works~\cite{MSTCN,Dai,BCN,GCN1,GCN2,ED-TCN}. The dimension of the I3D feature for each frame is 2048-d. The following three evaluation metrics are used to evaluate the performance: frame-wise accuracy(Acc.), segmental edit score (Edit), and segmental overlap F1 score with threshold $k$/100, denoted as F1@$k$.

\section{Experiments}
\subsection{Impact of position encoding and multi-head self-attention}
In terms of implementation details, our ASFormer has two differences compared to the vanilla Transformer. First, the position encoding is not applied for the input of both encoder and decoders. Second, vanilla Transformer usually adapts multi-head attention that concatenates the output from multiple single-head attention to enhance the feature transformation ability. Instead, we only use a single-head attention in each encoder/decoder block. We conduct the following two experiments to demonstrate the introduction of temporal convolution makes our model free from position encoding and multi-head version of self-attention.

The first experiment studies whether position encoding is still needed for ASFormer. The results on 50Salads dataset are shown in Table~\ref{pe}. For the temporal convolution, the position encoding is even counterproductive. When the position encoding is only applied for the encoder, the performance already drops a lot. Further, when the position encoding is applied for both encoder and decoders, the performance is even worse. The possible reason is that the temporal convolution already has the ability to model the relative positional relationships. The redundant absolute position encoding might be harmful to the temporal convolutions to learn the feature embedding. Thus, we drop the position encoding from our model. The second experiment studies the impact of multi-head self-attention. Table~\ref{multi-head} shows the comparison of multi-head self-attention with the different numbers of heads on 50Salads. For each head in the multi-head attention, the feature dimension is the same as the single-head attention. We can observe that the single head self-attention achieves comparable performance with the multiple ones. This is because the convolution operation has a similar form with the multi-head self-attention operation (\eg{ a 64-filters convolution operator also concatenates the output from 64 1-filter convolutions}). Considering the extra computation and memory budgets brought by the multi-head self-attention, we use the single-head self-attention by default.  

\begin{minipage}[t]{\textwidth}
\footnotesize
\setlength\tabcolsep{2.0pt}
 \begin{minipage}[t]{0.4\textwidth}
     \makeatletter\def\@captype{table}\makeatother\caption{Comparison of whether using position encoding on 50Salads.\\}
       \begin{tabular}{clcc}
			\hline
			~  & F1@\{10, 25, 50\} & Edit & Acc.
			\\
			\textit{w/o pe (ours)} & \textbf{85.1\; 83.4\; 76.0} & \textbf{79.6} & \textbf{85.6} \\
			\textit{+ pe encoder} &  80.2\; 78.0\; 70.0 & 73.0 & 83.6  \\
			\textit{\qquad+ pe decoder} &  78.1\; 76.7\; 69.4 & 72.0 & 84.1  \\
			\hline
		\end{tabular}
		\label{pe}
  \end{minipage} \qquad \qquad
  \begin{minipage}[t]{0.45\textwidth}
        \makeatletter\def\@captype{table}\makeatother\caption{Comparison of multi-head self-attention with different number of heads on 50Salads.}
         \begin{tabular}{clcc}
			\hline
			~  & F1@\{10, 25, 50\} & Edit & Acc.
			\\
			\textit{single head (ours)} &  85.1\; \textbf{83.4}\; 76.0 & \textbf{79.6} & 85.6 \\
			\textit{two heads} & 85.0\; 83.0\; 75.9 & 78.2 & \textbf{85.9} \\
			\textit{three heads} & \textbf{85.2}\; 83.2\; \textbf{76.8} & 78.1 & 85.3 \\
			\hline
		\end{tabular}
		\label{multi-head}
   \end{minipage}
\end{minipage}

\subsection{Effect of the local connective inductive bias}
In this section, we study the effect of bringing in local connective inductive bias with the temporal convolution. For comparison, we use an MLP as the feed-forward layer as the same as the vanilla Transformer. It worth noting that when removing the temporal convolution from our model, the position encoding is needed to bring in position information as the vanilla transformer does. In order to directly compare the the two solutions, we only use the encoder part for the experiments to avoid the impact of the refinement process. Results on the 50Salads dataset are shown in Table~\ref{mlp_tc}. We can observe that when MLP is used as the feed-forward layer, the performance drops greatly, especially on F1 scores and Edit score, which denotes that the model fails to model the temporal relationship among frames to produce smoother and consistent predictions. This also demonstrates that the fusion of neighboring local information plays an important role in the performance.

\begin{table}[h]
\caption{Comparison of MLP with position encoding and temporal convolution as the feed-forward layer on 50Salads. Results on the the \textbf{encoder} are reported.}
	\begin{center}
	\footnotesize
		\begin{tabular}{llcc}
			\hline
			~  & F1@\{10, 25, 50\} & Edit & Acc.
			\\
			\textit{MLP+PE} & 27.6\; 25.3\; 19.9  & 20.0 & 74.2  \\
			\textit{Conv (ours)} & \textbf{53.1\; 51.4\; 47.0} & \textbf{43.3} & \textbf{85.7} \\
			\hline
		\end{tabular}
	\end{center}
	
	\label{mlp_tc}
\end{table}

\vspace{-0.2cm}
\subsection{Effect of the hierarchical representation pattern}
The hierarchical representation pattern plays an important role in our ASFormer. To demonstrate its effectiveness, we conduct a non-hierarchical version by setting the size of the local window in all attention layers to 512, which is the largest window size in the last encoder/decoder block (out of memory if we directly set the size of the local window in all attention layers to the video length). The non-hierarchical version allows each self-attention layer to span their attention weights `freely'. Experiments on 50Salads dataset are shown in Table~\ref{pattern}. The performance of the non-hierarchical version drops significantly.

\begin{table}[t]    \caption{Comparison of hierarchical representation pattern and the non-hierarchical pattern (by setting the size of local window in all attention layers to 512) on 50Salads.}
	\begin{center}
	\footnotesize
		\begin{tabular}{llcc}
			\hline
			~  & F1@\{10, 25, 50\} & Edit & Acc.
			\\
			\textit{non-hierarchical} &  64.2\; 61.5\; 55.1 & 59.5 & 76.8  \\
			\textit{hierarchical (ours)} & \textbf{85.1\; 83.4\; 76.0} & \textbf{79.6} & \textbf{85.6} \\
			\hline
		\end{tabular}
	\end{center}
	
	\label{pattern}
\end{table}

\begin{figure}[h]
\vspace{-0.4cm}
	\centering
	\includegraphics[width=\textwidth]{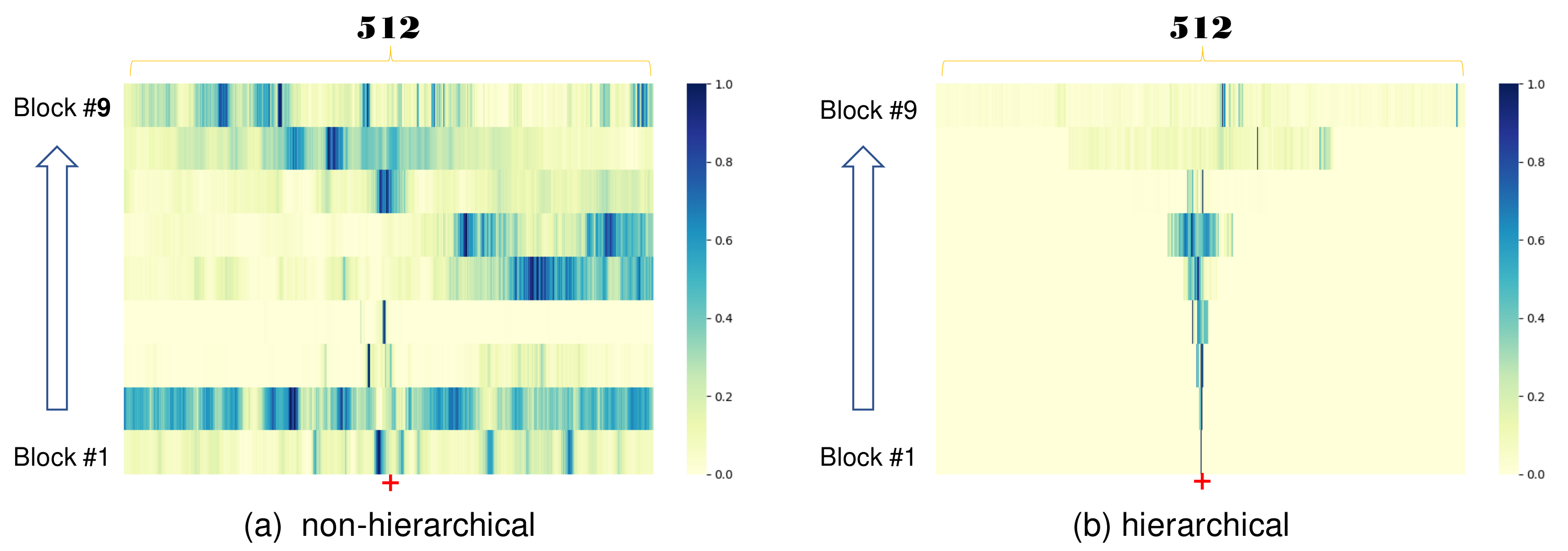}
	\caption{The visualization of attention weights for an anchor frame (red \textcolor{red}{+}) in each encoder block, more visualization can be found in the supplementary material. (a) the non-hierarchical (by setting the window size to 512 in all blocks). (b) With the hierarchical pattern.}
	\label{att_vis}
    \vspace{-0.3cm}
\end{figure}

To better understand why there is such a huge performance gap, we show some visualization results in Fig.~\ref{att_vis}. For an anchor query frame, we plot its attention weights in each self-attention layer in the encoder (The attention weights are normalized with min-max\footnote{$x = (x-min)/(max-min)$}.). Fig.~\ref{att_vis}(a) shows the non-hierarchical version, while Fig.~\ref{att_vis}(b) shows the case that with the pre-defined hierarchical pattern. We have two observations. First, for the high-level blocks (\eg{block \#9}) that has the same window size, there are much more activation (blue ribbon) in the non-hierarchical version after the min-max normalization. This means that the attention weights in many locations have close values and more similar to a trivial uniform distribution. In contrast, self-attention trained with the hierarchical pattern tend to focus on several meaningful locations. Secondly, the `freely' attention cannot automatically learn a hierarchical pattern from data. As shown in Fig.~\ref{att_vis}(a), the low-level blocks do not cast attention on its neighbors when the self-attention layer is not constrained. 

\subsection{Effect of the multiple decoders}
To demonstrate that our decoder takes advantage of temporal relations among multiple action segments for refinement, we conduct an ablation study about stacking the different numbers of decoders. Results on 50Salads dataset are shown in Table.~\ref{num_decoders}. The decoders largely boost the performance compared to the encoder, and we achieve the best results when stacking three decoders. To show the iterative refinement process, we further plot the predictions in the encoder and all decoders, as shown in the supplementary material. 

\subsection{Ablations of the number of blocks}
The number of blocks ${J}$ in the encoder/decoder is an important hyper-parameter, where more blocks bring larger receptive fields but also lead to higher memory costs as introduced in Sec.~\ref{encoder}. We conduct an ablation study about the different number of blocks in the encoder/decoder on 50salads dataset and report the peak value of the GPU memory cost for batch size 1 during training. Results are shown in Table~\ref{num_of_blocks}. We achieve higher performance with the increasing number of blocks until ${J}=9$. Using more than 9 blocks ($J=10$) slightly improves the frame-wise accuracy but does not increases the F1 scores. Meanwhile, the GPU memory cost increases dramatically with the large $J$. Thus, we choose ${J}=9$ for all our experiments.

\begin{minipage}[t]{\textwidth}
\footnotesize
\setlength\tabcolsep{2.0pt}
\begin{minipage}[t]{0.4\textwidth}
     \makeatletter\def\@captype{table}\makeatother\caption{\footnotesize{Comparison of stacking different number of decoders on 50Salads.}}
    	\begin{tabular}{clcc}
    		\hline
    		~  & F1@\{10, 25, 50\} & Edit & Acc.
    		\\
    		\textit{Encoder Only} & 53.1\; 51.4\; 47.0 & 43.3 & 85.7  \\
    		\hline
    		\textit{One Decoder} & 79.5\; 77.4\;, 71.6 & 71.5 & \textbf{86.8} \\
    		\textit{Two Decoders} & 83.9\; 82.8\; 76.8 & 76.7 & \textbf{86.8} \\
    		\textit{Three Decoders (ours)} & \textbf{85.1\; 83.4\; 76.0} & \textbf{79.6} & 85.6 \\
    		\textit{Four Decoders} & 84.0\; 82.2\; 75.8 & 76.5 & 84.3 \\
    		\hline
    	\end{tabular}
	\label{num_decoders}
\end{minipage} \qquad \qquad \qquad
\begin{minipage}[t]{0.4\textwidth}
    \makeatletter\def\@captype{table}\makeatother\caption{\footnotesize{Comparison of different number of blocks ${J}$ on 50Salads.\\}}
       \begin{tabular}{clccc}
			\hline
			${J}$  & F1@\{10, 25, 50\} & Edit & Acc. & GPU Mem.\\
			\hline
			\textit{7} &  82.9\; 81.5\; 74.0 & 76.0 & 84.5 & $\sim$2.1G \\
			\textit{8} &  84.4\; 82.8\; 75.4& 78.2 & 85.4 & $\sim$2.5G \\
			\textit{9} &  \textbf{85.1}\; 83.4\; 76.0 & \textbf{79.6} & 85.6 & $\sim$3.5G \\
			\textit{10} &  84.7\; \textbf{83.6\; 76.5} & 79.5 & \textbf{86.4} & $\sim$6.1G \\
			\hline
		\end{tabular}
		\label{num_of_blocks}
   \end{minipage}
\end{minipage}

\vspace{-0.3cm}
\subsection{Comparison with SOTA}
This section compares the proposed ASFormer to the state-of-the-art methods on three datasets: 50Salads, GTEA, and Breakfast datasets. The results are presented in Table~\ref{cmp_sota}. Our model achieves the state-of-the-art methods on the three datasets compared to previous work. The results on three metrics highlight the ability of our ASFormer to obtain accurate and smooth predictions.

\begin{table}[h]
	\begin{center}
	\caption{Comparison with the state-of-the-art on 50Salads, GTEA and the Breakfast dataset. \textbf{Bold} and \underline{underlined} denote the highest and the second value in each column.}
	\label{cmp_sota}

	\footnotesize
	\setlength\tabcolsep{3.5pt}
	\begin{tabular}{l|c|c|c|c|c|c|c|c|c}
		\hline
		 & \multicolumn{3}{c|}{\textbf{50Salads}} & \multicolumn{3}{c}{\textbf{GTEA}} & \multicolumn{3}{|c}{\textbf{Breakfast}} \\
		 \cline{2-10}
		 & F1@\{10,25,50\} & Edit & Acc. &  F1@\{10,25,50\} & Edit & Acc. & F1@\{10,25,50\} & Edit & Acc. \\
		 \hline
	    IDT+LM~\cite{IDT_LM} & 44.4\; 38.9\; 27.8& 45.8 & 48.7 & - & - &- & - & - &- \\
	    ST-CNN~\cite{stcnn} & 55.9\;  49.6\;  37.1 & 45.9& 59.4 & 58.7\; 54.4\; 41.9 & - & 60.6 & - & - & - \\
	    Bi-LSTM~\cite{bi-lstm}  & 62.6\; 58.3\; 47.0  & 55.6 & 55.7  & 66.5\; 59.0\; 43.6  & - & 55.5 &  & - & - \\
	    ED-TCN~\cite{ED-TCN}   & 68.0\; 63.9\; 52.6  & 59.8 & 64.7  & 72.2\; 69.3\; 56.0  & - & 64.0   & -  & - & 43.3     \\
	    HTK~\cite{HTK}  & - & - & - & - & - & - &  - & - & 50.7     \\
		TCFPN~\cite{TCFPN}  & - & - & - & - & - & - &  - & - & 52.0      \\
		SA-TCN~\cite{Dai}  & - & - & - & - & - & - &  - & - & 50.0      \\
		HTK(64) ~\cite{HTK(64)} & - & - & - & - & - & - &  - & - & 56.3      \\
	    TDRN~\cite{TDRN}      & 72.9\; 68.5\; 57.2  & 66.0 & 68.1 & 79.2\; 74.4\; 62.7  & 74.1 & 70.1 & - & - & - \\
	    SSA-GAN~\cite{SSA-GCN}    &74.9\; 71.7\; 67.0 & 69.8 & 73.3 & 80.6\; 79.1\; 74.2  & 76.0 & 74.4 & - & - & -\\
	    MS-TCN~\cite{MSTCN}     & 76.3\; 74.0\; 64.5  & 67.9 & 80.7 & 85.8\; 83.4\; 69.8  & 79.0 & 76.3 &  52.6\; 48.1\; 37.9    & 61.7 & 66.3 \\
	    DTGRM~\cite{GCN2}       & 79.1\; 75.9\; 66.1   & 72.0 & 80.0 & 87.8\; 86.6\; 72.9 & 83.0 & 77.6 & 68.7\; 61.9\; 46.6 & 68.9 & 68.3 \\
	    BCN~\cite{BCN}       & 82.3\; 81.3\; 74.0 & 74.3 & 84.4 & 88.5\; 87.1\; 77.3 & 84.4 & \textbf{79.8} & 68.7\; 65.5\; 55.0 & 66.2 & 70.4 \\
	    Gao \etal~\cite{Gao} & 80.3\; 78.0\; 69.8 & 73.4 & 82.2 & \underline{89.9}\; 87.3\; 75.8& 84.6 & 78.5 & \underline{74.9}\; \underline{69.0}\; 55.2 & \underline{73.3} & 70.7 \\
	    ASRF~\cite{bound2} & \underline{84.9}\; \textbf{83.5}\; \textbf{77.3} & \underline{79.3} & 84.5 & 89.4\; 87.8\; \textbf{79.8} & 83.7 & 77.3 & 74.3\; 68.9\; 56.1 & 72.4 & 67.6 \\

	    \hline
	    ASFormer & \textbf{85.1}\; \underline{83.4\;} \underline{76.0} & \textbf{79.6} & \textbf{85.6} & \textbf{90.1}\; \textbf{88.8}\; \underline{79.2} & \textbf{84.6} &  \underline{79.7} & \textbf{76.0\; 70.6\;} \textbf{57.4} & \textbf{ 75.0} & \textbf{73.5}\\
	    \hline
	    
	\end{tabular}
	\end{center}
	\vspace{-0.4cm}
\end{table}

\begin{minipage}[t]{\textwidth}
\footnotesize
\setlength\tabcolsep{2.0pt}
\begin{minipage}[t]{0.4\textwidth}
     \makeatletter\def\@captype{table}\makeatother\caption{\footnotesize{Comparison of ASRF(build upon \textit{MS-TCN}) and ASRF*(build upon our \textit{ASForme}r) on 50salads dataset.}}
       \begin{tabular}{clcc}
			\hline
			~  & F1@\{10, 25, 50\} & Edit & Acc.\\
			\hline
			ASRF\tiny{(\textit{MS-TCN})} &  84.9\; 83.5\; 77.3 & 79.3 & 84.5 \\
			ASRF*\tiny{(\textit{ASFormer})} &  \textbf{86.8\; 85.4\; 79.3} &  \textbf{81.9} & \textbf{85.9} \\
			\hline
		\end{tabular}
		\label{asrf_mstcn}
\end{minipage} \qquad \qquad
\begin{minipage}[t]{0.4\textwidth}
    \makeatletter\def\@captype{table}\makeatother\caption{\footnotesize{Comparison of ASFormer and MS-TCN on number of parameters, FLOPs and GPU memory cost on 50salads dataset.}}
    	\begin{tabular}{cccc}
			\hline
			~  & \#params ($M$) & FLOPs($G$) & GPU Mem.
			\\
			MS-TCN &  0.799 & 4.79 & $\sim$1.7G \\
			ASFormer & 1.134 & 6.80 & $\sim$3.5G \\
			
			\hline
		\end{tabular}
		\label{mstcn_compu}
   \end{minipage}
\end{minipage}
\vspace{0.3cm}

It is worth noting that, our ASFormer can serve as a strong backbone model and thus can be easily adapted by other works. To demonstrate that, we replace the original TCN-based backbone model MS-TCN~\cite{MSTCN} in ASRF~\cite{bound2} with our ASFormer. The new model, denoted as ASRF*, achieves even higher results on the 50salads dataset than the original ASRF~\cite{bound2}, as shown in Table~\ref{asrf_mstcn}. Besides the performance, we further compare with the popular TCN-based backbone MS-TCN~\cite{MSTCN} in terms of the computation cost, as shown in Table~\ref{mstcn_compu}. The additional computational burden of our ASFormer is acceptable with the modern chips. Thus, our ASFormer provides an alternative backbone choice for the community to build their models.

\section{Conclusion}
In this paper, we explore the application of Transformer-based models on action segmentation task. We propose three major concerns and respective solutions in our ASFormer. The superior results on three public dataset demonstrate the effectiveness of our method, and provide a new solution for community to address the action segmentation task. 

\noindent
\textbf{Acknowledgement.} This work was partially supported by the Natural Science Foundation of China under contracts 62088102. We also acknowledge High-Performance Computing Platform of Peking University for providing computational resources.

\bibliography{egbib}
\end{document}